\def\year{2022}\relax
\documentclass[letterpaper]{article} 
\usepackage{aaai22}  
\usepackage{times}  
\usepackage{helvet}  
\usepackage{courier}  
\usepackage[hyphens]{url}  
\usepackage{graphicx} 
\usepackage{natbib}  
\usepackage{caption} 
\DeclareCaptionStyle{ruled}{labelfont=normalfont,labelsep=colon,strut=off} 
\usepackage{algorithm}
\usepackage{algorithmic}

\usepackage{newfloat}
\usepackage{listings}
\usepackage{color,xcolor}
\usepackage{amsmath}
\usepackage{txfonts}
\usepackage{microtype}
\usepackage{graphicx}
\usepackage{subfigure}
\usepackage{booktabs} 
\usepackage{array}
\AtBeginDocument

\usepackage{microtype}
\usepackage{graphicx}
\usepackage{subfigure}
\usepackage{amssymb}
\usepackage{booktabs} 
\usepackage{multirow}
\usepackage{adjustbox}

\floatstyle{ruled}
\newfloat{listing}{tb}{lst}{}
\floatname{listing}{Listing}
\nocopyright
\title{Defeating Catastrophic Forgetting via Enhanced Orthogonal Weights Modification}
\author {
    Yanni Li\textsuperscript{\rm 1},
    Bing Liu\textsuperscript{\rm 2},
    Kaicheng Yao\textsuperscript{\rm 1}\equalcontrib, 
    Xiaoli Kou\textsuperscript{\rm 1}\equalcontrib,
    Pengfan Lv\textsuperscript{\rm 1}\equalcontrib,
    Yueshen Xu\textsuperscript{\rm 1},
    Jiangtao Cui \textsuperscript{\rm 1}
}
\affiliations {
    \textsuperscript{\rm 1} School of Computer Science and Technology, Xidian University\\
    \textsuperscript{\rm 2} Department of Computer Science, University of Illinois at Chicago\\
    yannili@mail.xidian.edu.cn,liub@uic.edu,\{kcyao,pflv\}@stu.xidian.edu.cn,\{xlkou,ysxu,cuijt\}@xidian.edu.cn.
}

\begin{document}

\maketitle

\begin{abstract}
The ability of neural networks (NNs) to learn and remember multiple tasks sequentially is facing tough challenges in achieving  general artificial intelligence due to their \textit{catastrophic forgetting} (CF) issues. Fortunately, the latest OWM (\underline{\textbf{O}}rthogonal \underline{\textbf{W}}eights \underline{\textbf{M}}odification) and other several \textit{continual learning} (CL) methods suggest some promising ways to overcome the CF issue. However, none of  existing CL methods explores the following three crucial questions for effectively overcoming the CF issue: that is, what knowledge does it  contribute to the effective weights modification of the NN during its sequential tasks learning? When the data distribution of a new learning task changes corresponding to the previous learned tasks, should a uniform/specific weight modification strategy be adopted or not? what is the upper bound of the learningable tasks sequentially for a given CL method? ect. To achieve this, in this paper, we first reveals the fact that of the weight gradient of a new learning task is determined by both the input space of the new task and the weight space of the previous learned tasks sequentially. On this observation and the recursive least square optimal method, we propose a new efficient and effective continual learning method EOWM via enhanced OWM. And we have theoretically and definitively given the upper bound of the learningable tasks sequentially of our EOWM. Extensive experiments conducted on the benchmarks demonstrate that our EOWM is effectiveness and outperform all of the state-of-the-art CL baselines.
\end{abstract}

\section{Introduction}
Current state-of-the-art deep neural networks (NNs) can be trained to impressive performance on a wide variety of individual tasks \cite{r56,r57,r59}. Learning multiple tasks in sequence, however, remains a substantial challenge in the NNs. When training on a new task, standard NNs will forget most of the information related to previously learned tasks, a phenomenon referred to as \textit{catastrophic forgetting} (CF)  \cite{r1, r41}. Without solving this problem, the NN is hard to adapt to \textit{lifelong or continual learning}(CL), which is crucial and fundamental for \textit{general artificial intelligence} (GAI).

\textbf{Problem Statement}: \textit{Given a sequence of supervised learning tasks $T = (T_1, T_2, ..., T_N)$, we want to learn them one by one in the given sequence such that the learning of each new task will not forget the models learned for the previous tasks} \cite{r41, r18}.

For the CL research, there have been two main directions over the past 30 years, i.e., one from biological aspects, and the other from deep learning\cite{r16}. In this study, we focus on the latter. In recent years, many CL methods have been proposed to lessen the effect of CF \cite{r18, r16}, e.g., learning without forgetting (LWF) \cite{r43}, elastic weight consolidation (EWC) \cite{r2}, synaptic intelligence (SI)\cite{r51}, dynamically expandable network (DEN) \cite{r42}, progressive neural networks (PNN) \cite{r52}, averaged gradient episodic memory (AGEM) \cite{r30}, parameter generation and model adaptation (PGMA) \cite{r18}, learn to grow (LTG) \cite{r23}, maximally interfered retrieval(MIR) \cite{r55}, online fast adaptation and knowledge accumulation (OSAKA) \cite{r44}, etc. Despite the advantages of the above CL methods, however, none of  the above  existing CL methods explores the following three crucial questions for effectively overcoming the CF issue: such as, what knowledge does it  contribute to the effective weights modification of the NN during its sequential tasks learning? When the data distribution of a new learning task changes corresponding to the previous learned tasks, should a uniform/specific weight modification strategy be adopted or not? what is the upper bound of the learningable tasks sequentially for a given CL method? 
According to whether task identity is provided and whether it must be inferred during test, there are mainly three CL scenarios \cite{r45, r17}, i.e., task incremental learning, domain incremental learning and \textit{class incremental learning} (CIL). CIL is the most challenging scenario in which the classes of each task are assumption of joint or disjoint and the model is trained to distinguish classes of all tasks with a shared output layer, namely single-head.
Among the existing CL algorithms, it is worth mentioning that OWM (\underline{\textbf{O}}rthogonal \underline{\textbf{W}}eights \underline{\textbf{M}}odification) algorithm \cite{r8} and its few  variants \cite{r7, r12, r14, r15}, heareafter abbreviated as OWM algorithms, emerged in recent years are suggested to be a promising solution to the CIL scenario as their excellent theory, interpretability and performance \cite{r45,r17,r8}. In the OWM algorithms, a good OWM projector $P_{OWM}$, which uses to find a gradient orthogonal direction, is the key of its method performance. Although the above latest few OWM algorithms have achieved gratifying results, however, we found that the existing OWM algorithms have following weaknesses: 1) none of their $P_{OWM}$ adopt all the necessary knowledge, 2) they make an unpractical assumption of being disjoint between classes of sequential tasks, and 3) they have considerably less attention to the rigorous evaluation of the algorithms. 
In this study, aiming to achieve all of the above weaknesseswe of OWM and above representative CL algorithms, we propose a new improved enhanced OWM algorithm, namely EOWM (\underline{\textbf{E}}nhanced \underline{\textbf{O}}rthogonal \underline{\textbf{W}}eights \underline{\textbf{M}}odification) . Our contributions can be summarized as follows:
\begin{itemize}
    \item We theoretically study the projector $P_{OWM}$ of the existing OWM algorithms, andwe first reveals the fact that of the weight gradient of a new learning task is determined by both the input space of the new task and the weight space of the previous learned tasks sequentially. On the basis, we propose an enhanced projector $P_{EOWM}$, which can achieve all the above weaknesses of exsiting OWM and other CL algorithms. 
    
    
     \item Based on our proposed $P_{EOWM}$ and the recursive least square optimal method, we present a new efficient enhanced OWM algorithm EOWM followed by introducing a more reasonable metric to address the weakness of considerably less attention to the rigorous evaluation of the OWM algorithms.
    
     \item Extensive experiments conducted on benchmark datasets show that our proposed algorithm  EOWM achieves a new SOTA  (\underline{S}tate-\underline{O}f-\underline{T}he-\underline{A}rt) results compared with some competitive baselines.
\end{itemize}

This paper is organized as follows. Section 2 first gives some preliminaries, and overviews related work. Section 3 details our theoretical study to the projector $P_{OWM}$ of the OWM methods followed by our proposed algorithm EOWM. Section 4 reports experimental results on benchmark datasets to evaluate our method. Finally, Section 5 concludes this study.

\section{Related Work}
In this study, as we focus on a new and enhanced OWM algorithm, which is expected to effectively overcome the above weaknesses of the existing CL and OWM algorithms, we first outline the OWM methods, and then overview the latest related OWM algorithms.

\subsection{Preliminaries}
\textbf{OWM method}: It is also called \textit{subspace methods} \cite{r7,r8}. Its basic strategy used in these  algorithms is that it retains previously learned knowledge by keeping the old input-output mappings that NNs induce fixed. To meet this goal, the gradients are projected to the subspace that is orthogonal to the inputs of past tasks.
More formally, in OWM, \textit{a projector $P_{OWM}$} used to find the orthogonal direction to the input space is defined as \cite{r8} 

\begin{equation}
P_{OWM} = I-A_{n-1} \left ( A_{n-1}^TA_{n-1}+\alpha I\right)^{-1}A_{n-1}^T\label{eq_1} 
\end{equation}
where $A_{n-1}$ consists of all $n-1$ previously trained input vectors as its columns $A_{n-1}$ = $(\boldsymbol{x_1},..., \boldsymbol{x_{n-1}})$, and $I$ is a $n-1$ order unit matrix multiplied with a relatively small constant $\alpha$. For the n-th new learning task, the learning-induced weights modification of a NN is then updated by the weight gradient, which is calculated according to the standard back-propagation and modified by the projector $P_{OWM}$ \textit{It is worth noting that a well-designed projector $P_{OWM}$ is the vital to a  effective and efficient OWM algorithm}.

\subsection{Related work}


Among a few OWM algrithms, OWM method \cite{r8} is the pioneer of this kind of algorithms. Inspired by the role of the prefrontal cortex (PFC) in mediating context-dependent processing in the primate brain, the OWM first proposed a novel orthogonal weights modification method with the addition of a PFC-like module, that enables CNNs to continually learn different mapping rules in a context-dependent way  without interference leading to reach a human level ability in online and continual learning. 

CAB \cite{r7} proposed a variant of the back-propagation algorithm based on OWM basic principle, i.e., conceptor-aided backprop (CAB), in which gradients are shielded by conceptors against degradation of previously learned tasks. On the benchmark datasets CAB outperforms its baselines for coping with catastrophic interference.

Inspired by the first OWM method \cite{r8}, \cite{r12} proposed an improve OWM method, namely OWM+GFR, with the strategy of directly generating and replaying features. its empirical results on image and text datasets showed that the OWM+GFR can improve OWM \cite{r8} consistently by a significant margin while conventional generative replay always resulted in a negative effect, and beats a SOTA generative replay method \cite{r18}. 

From the parameter space perspective, OGD \cite{r14} and OGD+ \cite{r15} studied an approach to restrict the direction of the gradient updates to avoid forgetting previous-learned tasks, which accomplished its goal by projecting the gradients from new tasks onto a subspace where the neural network output on previous tasks do not change and the projected gradient is still in a useful direction. Moreover, OGD+ proved that OGD was robust to CF then derived the first generalization bound for SGD (\underline{S}tochastic \underline{G}radient \underline{D}escent) \cite{r40} and OGD for CL. 

Though the above few pioneering OWM algorithms claimed that they were the most promising solution to the CF problem, we suggest that they have three serious weaknesses (see Para. 5 of Introduction for details). Different from all of the above OWM methods, based on our findings we proposed a new enhanced OWM method, namely EOWM, to achieve all the above weaknesses of the existing OWM and other CL methods.

\section{The Proposed Enhanced OWM}
\label{sect:theproposedenhancedowm}

\subsection{Basic Concepts}
For clarity, based on optimization theory\cite{r13},  we first introduce following basic concepts and notations.

\textbf{Definition 1: Weight Space}. For a NN, let $W_{i}$ be the weight matrix of a layer learning the $i$-th task $T_{i}$ sequentially. And define $\Omega_{n-1}$ as the weight space of learned previous $n-1$ tasks, which consists of all $n-1$ previously weight matrices as its columns, i.e., $\Omega_{n-1}=(W_1,W_2,...,W_{n-1})$. And denote  $\Omega_{n-1}^\perp$ as the orthogonal subspace of $\Omega_{n-1}$.

Given a NN, its weight matrix $W_n$ for the learning task $T_n$ can be calculated by the gradient descent method \cite{r58} as follows. 
\begin{equation}
W_n = W_{n-1} - \lambda \Delta W_{n} \label{eq_2}
\end{equation}
where $\lambda$ is the learning rate and $\Delta W_n$ corresponds to the weight gradient for the task $T_{i}$.

\textbf{Definition 2: Projections}. Let $\Delta W_n^{\Omega_{n-1}}$($\Delta W_{n}^{\Omega_{n-1}^\perp}$) be the projections of $\Delta W_n$ on $\Omega_{n-1}$ ($\Omega_{n-1}^\perp$). From the basic definition of projection\cite{r13}, we have 
\begin{equation}
\label{eq_3}
    \begin{cases}
    \Delta W_{n}=\Delta W_{n}^{\Omega_{n-1}}+\Delta W_{n}^{\Omega_{n-1}^{\perp }} \ \\
    s.t.\  \Delta W_{n}^{\Omega_{n-1}} \in \Omega_{n-1} ,\  \Delta W_{n}^{\Omega_{n-1}^{\perp }} \in \Omega_{n-1}^{\perp }
    \end{cases}
\end{equation}

\textbf{Definition 3: Projection operator}. Let $Q$ and $Q_{ort}$ be the projection operators that project $\Delta W_{n}$ to $\Omega_{n-1}$ and  $\Omega_{n-1}^{\perp}$, respectively. Based on the basic definition of the projection operator\cite{r11,r13}, $Q$ and $Q_{ort}$ can be computed as follows. 
\vspace{-2mm}
\begin{equation}
\small{
\label{eq_4}
\begin{cases}
Q = \Omega_{n-1}\left(\Omega_{n-1}^T \Omega_{n-1}+\beta I\right)^{-1}\Omega_{n-1}^T \ \\
Q_{ort} = I- \Omega_{n-1}\left(\Omega_{n-1}^T \Omega_{n-1}+\beta I\right)^{-1}\Omega_{n-1}^T  \
\end{cases}
}
\end{equation}
where $\beta$ is a very small positive  empirical constant.

On the above basis, following Eq.\ref{eq_5} can be deduced.


\begin{equation}
\small{
\label{eq_5}
\begin{cases}
\Delta W_{n}^{\Omega_{n-1}}=\Delta W_{n}Q \ \\
\Delta W_{n}^{\Omega_{n-1}^{\perp }}=\Delta W_{n}Q_{ort}\ \\
\end{cases}
}
\end{equation}

Based on Eqs.\ref{eq_3} to \ref{eq_5}, following Theorems can be obtained.

\textbf{Theorem 1: }$\Delta W_{n}^{\Omega_{n-1}}$ and $\Delta W_{n}^{\Omega_{n-1}^\perp}$ only contain the knowledge of the previous learned tasks $(T_1,T_2,...,T_{n-1})$ and the current learning task $T_n$, respectively.

\vspace{1mm}
\textbf{\textit{Proof}:} As $\Delta W_{n}^{\Omega_{n-1}} \in \Omega_{n-1}$ and $\Delta W_{n}^{\Omega_{n-1}^{\perp }} \in \Omega_{n-1}^{\perp }$, $\Delta W_{n}^{\Omega_{n-1}}$ only contains the knowledge of previous learned tasks. $\Omega_{n-1}^{\perp }$ is orthogonal to $\Omega_{n-1}$, so $\Delta W_{n}^{\Omega_{n-1}^{\perp}}$ only contains the knowledge of the new task $T_n$. $\hfill\blacksquare$

\vspace{1mm}
\textbf{Theorem 2: }When a NN learns a new task $T_n$, both the two projections $\Delta W_{n}^{\Omega_{n-1}}$ and $\Delta W_{n}^{\Omega_{n-1}^{\perp }}$ of the weight gradient $\Delta W_{n}$ of $T_n$ will disturb the weight space $\Omega_{n-1}$ of the previous learned tasks $(T_1,T_2,...,T_{n-1})$, and $\Omega_{n-1}$ would be transformed into a new weight space $\Omega_{n}$ as follows.
\begin{equation}
\small{
\begin{aligned}
\label{eq_6}
\Omega _n =\left[\Omega_{n-1},-\lambda \left(\Delta W_{n}^{\Omega_{n-1}}+ \Delta W_{n}^{\Omega_{n-1}^{\perp}}\right)\right]
\end{aligned}}
\end{equation}
where  $\Delta W_{n}^{\Omega_{n-1}} \in \Omega_{n-1}$, and $ \Delta W_{n}^{\Omega_{n-1}^{\perp }} \not\in \Omega_{n-1}$.\\

\vspace{1mm}
\textbf{\textit{Proof}:} After the new task $T_n$ with the weight matrix $W_n$ is trained, its weight space  has changed  as $\Omega_{n} = \left [ W_1,W_2,...,W_{n-1},W_n\right]$. 

With Eq.\ref{eq_2},  $\Omega_{n}=\left[W_1,W_2,...,W_{n-1},\left(W_{n-1}-\lambda \Delta W_{n} \right)\right]$.
Following the equivalence of elementary transformation of matrix and Eq.\ref{eq_3}, we have 
\begin{equation}
\ \ \ \ \ \ \ \ \ \ \ \ \ \ \ \ \ \ \ \ \ \ \ \ \ \ \ \ \ \
\small{
\begin{aligned}
    \Omega_{n}&= \left[W_1,W_2,...,W_{n-1},-\lambda \Delta W_{n}\right]\\
    &=\left[W_1,W_2,...,W_{n-1},-\lambda \left(\Delta W_{n}^{\Omega_{n-1}}+\Delta W_{n}^{\Omega_{n-1}^{\perp}}\right)\right] \\
&=\left[\Omega_{n-1},-\lambda \left(\Delta W_{n}^{\Omega_{n-1}}+ \Delta W_{n}^{\Omega_{n-1}^{\perp}}\right)\right]\ \ \ \ \ \ \ \ \ \ \ \ \ \ \ \ \ \ \  \hfill\blacksquare   \nonumber 
\end{aligned}
} 
\end{equation}

 In particular, the above Theorems and Eq.\ref{eq_6}  clearly reveals the fact that \textit{the new weight space for the new task $T_n$ consists of two parts, where one is the previous learned weight space $\Omega_{n-1}$, and another is the influence of old knowledge ($\Delta W_{n}^{\Omega_{n-1}}$) and new knowledge ($\Delta W_{n}^{\Omega_{n-1}^{\perp}}$) on $\Omega_{n-1}$.}
\subsection{ An Enhanced Projection Operator $P_{EOWM}$}
\begin{figure*}[h]
    \centering
    \subfigure[The classes in task $T_{A}$ and task $T_{B}$ intersect, denoted as $T_{A} \simeq T_{B}$.]{
        \includegraphics[scale=0.9]{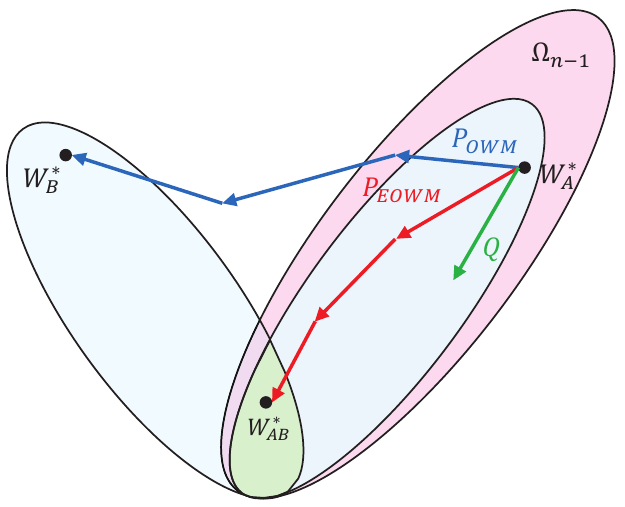}
    }\hspace{20mm}
    \subfigure[The classes in task $T_{A}$ and task $T_{B}$ disjoint, denoted as $T_{A} \not\simeq T_{B}$.]{
        \includegraphics[scale=0.9]{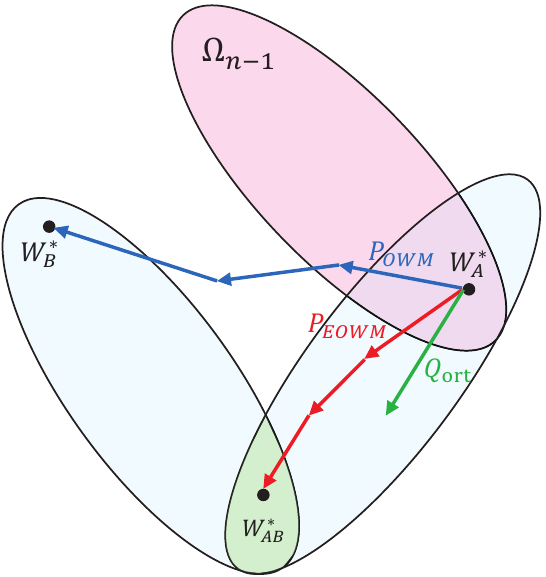}
    }
    \caption{The schematic diagram of moving trajectories of $P_{OWM}$ and proposed $P_{EOWM}$, where $W_A^*$ (pink area), $W_B^*$ (blue area) and $W_{AB}^*$ (green area) represent the optimal weights about task A, task B, and both task A and task B, respectively. Both Figs.(a) and (b) clearly indicate that $P_{EOWM}$ always progresses towards optimal direction $W_{AB}^*$ of both the previous learned task A and the new learning task B, while $P_{OWM}$ does towards the optimal direction $W_B^*$ of the new learning task B.}
    \label{fig_1}
\end{figure*}
Equipped with the above revealed important fact, we aim to seek an enhanced projecting operator denoted as $P_{EOWM}$ to overcome the existing OWM and other CL methods' weaknesses. For clarity, we first introduce Def.5 followed by our proposed $P_{EOWM}$.
\textbf{Definition 5: } For simplicity, let the class label set of all the previous learned/trained tasks $T_{prev}$ as $C_{prev}$, while the class label set of new learned/trained task $T_{new}$ as $C_{new}$. If $C_{prev} \cap C_{new} \neq \varnothing $, we define that $T_{prev}$ is similar to or joints $T_{new}$ denoted as $T_{prev} \simeq T_{new}$. Otherwise, both  $T_{pewv}$ and $T_{new}$ are dissimilar, namely disjoint denoted as $T_{prev} \not\simeq T_{new}$.

\textbf{An Enhanced Projection Operator $P_{EOWM}$: }In practice, the classes of sequentially learning tasks may be joint or disjoint. Considering the both situations, we calculate the $P_{EOWM}$ as follows:

1) if $T_{prev} \simeq T_{new}$, The weight space $\Omega_{n-1}$ of $T_{prev}$ most likely to contains optimal weights of the new task $T_{new}$. Based on the Def. 3 and Theorem 2, the correction matrix of the weight gradient of $T_{new}$, i.e., $P_{EOWM}$, should increase a gradient change over $\Omega_{n-1}$;  Otherwise,  

2) the optimal network weights to the new  task $T_{new}$ on $\Omega_{n-1}^{\perp }$  should be ensured and enhanced. As $\Omega_{n-1}^{\perp }$ and $\Omega_{n-1}$ are orthogonal, the learning of the optimal network weights of  $T_{new}$  will not affect the original network weights of the previous tasks $T_{prev}$. 

Fig.\ref{fig_1} illustrates the core idea of our $P_{EOWM}$, that is, \textit{$P_{EOWM}$ will always guide the weights gradient towards the optimal direction for previous learned tasks and new learning task}. In short, our $P_{EOWM}$ can be formalized as follows:
\begin{equation}
    \label{eq_7}
    P_{EOWM}=
    \begin{cases}
    P_{OWM}(c_1I+c_2Q),\ T_{prev} \simeq T_{new} \ \\
    P_{OWM}(c_1I+c_2Q_{ort}),\ T_{prev} \not\simeq T_{new} \
    \end{cases}
\end{equation}
where $c_1$ and $c_2$ are positive empirical coefficients less than 1, and $c_1+c_2=1$. $Q$ and $Q_{ort}$ are the projection operators that project $\Delta W_{n}$ to $\Omega_{n-1}$ and $\Omega_{n-1}^{\perp}$, respectively (see Def.3 and Eq.\ref{eq_4}).

Let the correction matrix of the weight gradient for new task $T_{n}$ be $P_{EOWM}^n$. $P_{EOWM}^n$ is used to modify its weight gradient $\Delta W_{n}$, According to Eq.\ref{eq_2}, the weight of a new task $T_{n}$ can be updated by following Eq.\ref{eq_8}.

\begin{equation}
\footnotesize{
\label{eq_8}
W_n= W_{n-1}-\lambda \Delta W_{n}P_{EOWM}^{n}\\
}
\end{equation}

\textbf{The efficient iterative calculation of $P_{EOWM}$: }To avoid the heavy computation cost of large-scale matrix operations in Eqs.\ref{eq_7} and \ref{eq_8}, Based on the Recursive Least Square (RLS) algorithm\cite{r10}, an  efficient iterative calculation of $P_{EOWM}$ is carefully designed, which is shown in Theorem 3 for details.. 

\textbf{Theorem 3: } $P_{EOWM}$ can be calculated efficiently and iteratively by Eq.\ref{eq_9} (proof shown in Appedix A).
\begin{equation} 
    \label{eq_9}
    P_{EOWM}^{n+1}=
    \begin{cases}
    P_{OWM}^{n+1}(c_1I+c_2Q^{n+1}),\ T_{prev} \simeq T_{n+1} \ \\
    P_{OWM}^{n+1}(c_1I+c_2Q_{ort}^{n+1}),\ T_{prev} \not\simeq T_{n+1} \
    \end{cases}
\end{equation}
where $P_{OWM}^{n+1} = P_{OWM}^{n}  - \boldsymbol{\kappa}_x\bar{\textbf{x}}_n^TP_{OWM}^{n}$, and   $\boldsymbol{\kappa}_x=\left( \alpha+\bar{\textbf{x}}_n^TP_{OWM}^{n}\bar{\textbf{x}}_n\right) ^{-1}P_{OWM}^{n}\bar{\textbf{x}}_n $. $\bar{\textbf{x}}_n$ corresponds to the mean value of task $T_n$'s input, and $\alpha$ is a small constant same as in Eq.\ref{eq_2}, respectively.
The iterative equation of $Q_{ort}$ and $Q$ is given as follows.
\begin{equation}
\label{eq_10}
\begin{cases}
Q_{ort}^{n+1} &= Q_{ort}^{n} - \boldsymbol{\kappa}_{W}  \overline{\textbf{W}}_{n}^TQ_{ort}^{n} \ \\ 
Q^{n+1} &= I-Q_{ort}^n \ 
\end{cases}
\end{equation}
where $\boldsymbol{\kappa}_{W}=\left(\beta + \overline{\textbf{W}}_{n}^{T}Q_{ort}^{n}\overline{\textbf{W}}_{n} \right)^{-1}Q_{ort}^{n}\overline{\textbf{W}}_{n}$. $\overline{\textbf{W}}_{n}$ and $\beta$ are the
mean value of network weights after task $T_n$ learned and
the small constant in Eq.\ref{eq_4}, respectively.

\textit{Theorem 3 clearly indicates that the iterative computation of $P_{EOWM}$ only need the input of current learning task $T_n$ and its weight matrix $W_n$, which greatly reduces the storage space and speedup the calculation.}

\subsection{The proposed EOWM Algorithm}
\begin{figure}[!tbh]
\vskip 0.0in
\begin{center}
\centerline{\includegraphics[scale=0.85]{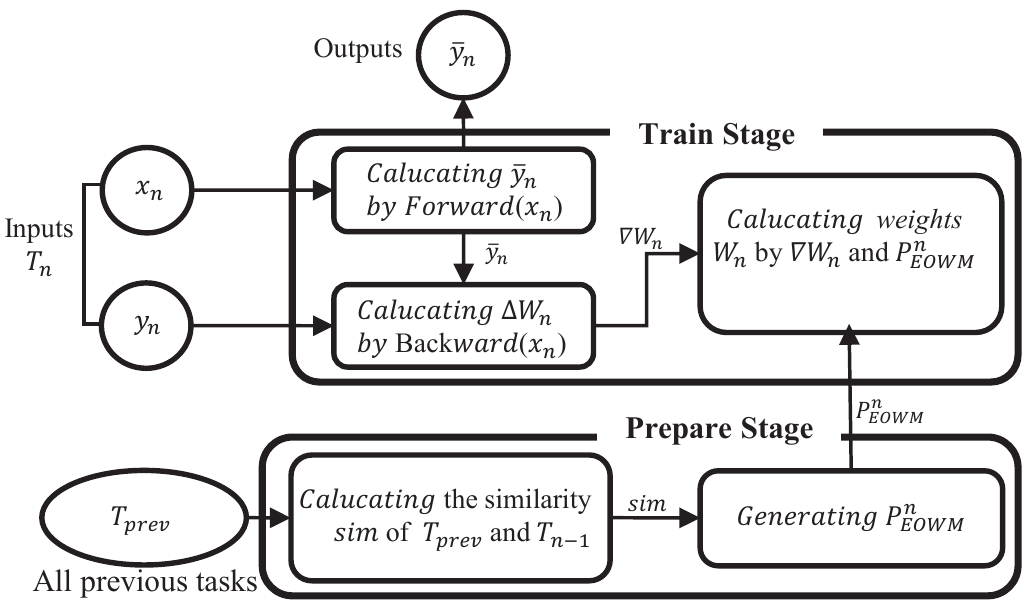}}
\caption{The training framework of our EOWM algorithm. }
\label{fig_2}
\end{center}
\vskip -0.2in
\end{figure}

On the above basis, we present a novel enhanced OWM algorithm, namely EOWM.  Its algorithm pseudo-code and training framework are shown in  Alg.\ref{algo1} and  Fig.\ref{fig_2}, respectively.

\begin{algorithm}[h]
\caption{EOWM algorithm}
\label{algo1}
\textbf{Input:} The sequential tasks $\{T_1,T_2,...,T_N\}$\\
\textbf{Parameter:} Regular terms $\alpha$ and $\beta$ of projection operator; Learning rate $\lambda$\\
\textbf{Output:} All tasks' predicted labels \textbf{$\bar{y}$}
\begin{algorithmic}[1] 
\STATE Randomly initialize the neural network weights $W$, $P_{OWM} \leftarrow I/\alpha$, $Q_{ort}\leftarrow I/\beta$, $T_{prev} \leftarrow \text{\o} $, \textbf{$\bar{y}$} $ \leftarrow \text{\o} $;
\WHILE{each $j \in [1,$  $N]$}
\STATE \textbf{$\bar{y}_{j}$} $\leftarrow \textbf{\textit{Forward}}(x_{j})$; \footnotesize{//\textbf{$\bar{y}_j$} is the predicted label for $T_j$.}
\STATE $\Delta W_{j} \leftarrow \textbf{\textit{Backward}}(\textbf{\textit{Loss}}(y_{j}$, \textbf{$\bar{y}_{j}$))};
\IF{j = 1}
\STATE Calculate $W_{j}$ by Eq.{\ref{eq_2}};
\ELSE
\STATE Calculate $W_{j}$ by Eq.{\ref{eq_8}};\footnotesize{// Update weights by  $P_{EOWM}$.}
\ENDIF
\STATE Calculate $P_{OWM}$ by Eq. \ref{eq_9};
\STATE Calculate $Q_{ort}$  and $Q$ by Eq. \ref{eq_10};
\STATE $sim=\textbf{\textit{Similarity}}(T_{prev},T_j)$; 
\STATE Calculate empirical coefficients $c_1$ and $c_2$ with $sim$ 
\STATE Calculate $P_{EOWM}$ by Eq. \ref{eq_9};\footnotesize{// Generate $P_{EOWM}$.}
\STATE $T_{prev} \leftarrow T_{prev} \cup T_j$;
\STATE \textbf{$\bar{y}$} $\leftarrow$ \textbf{$\bar{y}$} $\cup$ \textbf{$\bar{y_j}$};
\ENDWHILE
\STATE \textbf{return $\bar{y}$}.\\
\end{algorithmic}
\end{algorithm}

The proposed EOWM algorithm consists of three functional modules: 1) initialization  (shown in line 1 in Alg.1), 2) updating Weights of the $NN$ by $P_{EOWM}$ layer-by-layer  (shown in lines 3 to 9 in Alg.1), with the forward propagation  algorithm (\textbf{\textit{Forward}}(·)), back propagation algorithm \textbf{\textit{Backward}}(·), and Loss function (\textbf{\textit{Loss}}(·)), respectively. Note that the weights updating for the Task $T_{1}$ is calculate by Eq.\ref{eq_2}, while for the other tasks $T_{2}...T_{N}$ by Eq.\ref{eq_8}, and 3) generating final weights of the $NN$ by Eqs.\ref{eq_9} and \ref{eq_10} with similarity function (\textbf{\textit{Similarity}}(·)) (shown in lines 10 to 16 in Alg.1).

The training pipeline of the EOWM is shown in Fig.\ref{fig_2}. In training for task $T_{n}$, the correction matrix $P_{EOWM}^{n}$ is firstly calculated based on $P_{EOWM}^{n-1}$ of the previous tasks $T_{prev}$ ($T_{1}...T_{n-1}$). Then the class labels of the task $T_{n}$ are predicted with forward propagation algorithm \textbf{\textit{Forward}}(·). Finally, the training of current task $T_{n}$ is completed with the back propagation algorithm \textbf{\textit{Backward}}(·).

Especially noteworthy is that our proposed EOWM algorithm also has the following two unique properties (shown in Theorems 4 and 5) by our theoretical analysis and extensive experiments.

\textbf{Theorem 4: }The time complexity of our EOWM algorithm is  $O(N_c^3N_n)$, where $N_c$ and $N_n$ are the numbers of the columns of the weight and neurons of a layer in the NN, respectively (proof shown in Appedix B).

As the capacity of one layer of the $NN$ can be measured by the rank of $P_{EOWM} $, we can use the rank of $P_{EOWM} $, denoted as $rank(P_{EOWM})$, for the number of learnable different tasks\cite{r8}.

\textbf{Theorem 5 :} The upper bound of the minimum number of learnable tasks of our proposed EOWM is $rank(P_{EOWM})$, which is fomulated as follows. 
\begin{equation} 
\footnotesize{
    rank(P_{EOWM})<=
    \begin{cases}
    Rank_{OWM}+rank(P_{OWM}Q),\ T_{prev} \simeq T_{n+1} \nonumber\ \\
    Rank_{OWM}+rank(P_{OWM}Q_{ort}),\ T_{prev} \not\simeq T_{n+1} \nonumber\
    \end{cases}
    }
\end{equation}
where $Rank_{OWM}$ is the maximum rank of $P_{OWM}$ when $rank(P_{OWM}Q)>0$ and $rank(P_{OWM}Q_{ort})>0$ (proof shown in Appedix C).

\section{Experiments}
\subsection{Datasets and Experimental Settings}
We evaluate our EOWM and compare with some competitive baselines using three image datasets.
\vspace{-0.1cm}
\subsubsection{Datasets:}1) MNIST\cite{r53}: this dataset consists of 70,000 28$\times$28 black-and-white images of handwritten digits from 0 to 9. We use 60,000/3000/7000 images for training/validation/testing respectively. 2) CIFAR-10\cite{r54}: this dataset consists of 60,000 32$\times$32 color images of 10 classes, with 6000 images per class. We use 50,000/3000/7000 images for training/validation/testing respectively. 3) CIFAR-100\cite{r54}: this dataset consists of 60,000 32$\times$32 color images of 100 classes, with 600 images per class. We use 50,000/3000/7000 images for training/validation/testing respectively.
\vspace{-0.1cm}
\subsubsection{Data Preparation:}To simulate sequential learning, we adopt the same two data processing methods as in \cite{r60}, named disjoint and shuffled.

1) Shuffled (corresponding to similar tasks). We shuffle the input pixels of an image with a fixed random permutation to construct similar tasks. As shuffling real-world data will have a huge impact on the image, we only shuffle images in MNIST to construct Shuffled MNIST. We create three experimental settings of 3 tasks, 10 tasks and 20 tasks. In these cases, the dataset for the first task is the original dataset and the rest datasets are constructed through shuffling. Each task has 10 classes from 0 to 9.

2) Disjoint(corresponding to dissimilar tasks). In this case, we divide each dataset into several subsets of classes. Each subset represents one task. For example, we divide the CIFAR-10 dataset into two tasks. The first task consists of five classes $\{0, 1, 2, 3, 4\}$ and the second task consists of the remaining classes $\{5, 6, 7, 8, 9\}$. The $NN$ learns the two tasks sequentially and regards the two tasks together as 10-class classification. To study more tasks in test, on CIFAR-100 dataset, four experimental settings are created, which are 2 tasks, 5 tasks, 10 tasks and 20 tasks,respectively. Each task has the same number of classes.
\vspace{-0.1cm}
\subsubsection{Baselines:}To extensively evaluate our EOWM, we use seven state-of-the-art CL algorithms as our baselines: i.e., 1) EWC (\underline{E}lastic \underline{W}eight \underline{C}onsolidation) \cite{r2} is a typical regularization CL algorithms. 2) PGMA (\underline{P}arameter \underline{G}eneration and \underline{M}odel \underline{A}daptation) \cite{r18} is a competitive generative replay method that integrates generating network parameters and generating feature replay. 3) OGD (\underline{O}rthogonal \underline{G}radient \underline{D}escent) \cite{r14} is a representative  gradient projection method, which is robust to forgetting to an arbitrary number of tasks under an infinite memory. 4) OGD+ (\underline{O}rthogonal \underline{G}radient \underline{D}escent Plus) \cite{r15} is an extended method from OGD with more robustness than OGD. 5) OWM (\underline{O}rthogonal \underline{W}eights \underline{M}odification) \cite{r8} is the 
pioneer of the orthogonal gradient projection method. 6) ER-MIR and Hy-MIR\cite{r55} are two typical replaying methods, which select the most effective previous task data for replay.

All experiments are performed on a server with Intel(R) Xeon(R) Gold 5115 CPU@ 2.40GHz, 97GB RAM, NVIDIA Tesla P40 22GB GPU. Our EOWM  is implemented and tested using Python 3.6 and PyTorch 0.4.1. For the comparison fairness, we faithfully follow the running environments and hyper-parameters setting described in their original papers for all baselines in this paper.
\vspace{-0.1cm}
\subsubsection{Training Details:}To fairly compare, for similar tasks, our proposed EWOM algorithm uses the same classifier as all the above baselines. That is, a multilayer perception is adopted as the classifier, which consists of two fully connected layers with 100 neural cells as its hidden layer followed by a softmax layer. For dissimilar tasks, both the EOWM and baselines set a classifier same setting as \cite{r8}. That is, They consists of three layers CNN with 64, 128, and 256 2 x 2 filters, respectively, which have three fully connected layers with 1000 neural cells in their each hidden layer. To improve the ability of the NNs to extract features, the number of filters in NNs on CIFAR-100 dataset is doubled, that is, three NNs with 128, 256, and 512 2 x 2 filters,  respectively. The setting of fully connected layers is consistent with CIFAR-10 dataset. The hyper-parameters settings of our EOWM are  shown in Table. \ref{tab4_1}.

\begin{table}[h]
\setlength{\abovecaptionskip}{5.5pt}\centering
\caption{Hyper-parameters settings of proposed EOWM}\label{tab4_1}
\resizebox{0.4\textwidth}{12mm}{
\begin{tabular}{|c|ccc|}
\hline
Hyper-parameters & Shuffled MNIST & CIFAR-10 & CIFAR-100 \\ \hline
Epochs each task & 30 & 20 & 30 \\
Batch size & 100 & 100 & 100 \\
Learning rate & (3, 3.5) & 0.1 & 0.15 \\
$r$ & 0.99 & 0.99 & 0.99 \\
$\alpha$ & (0,1) & (0,1) & (0,1) \\
$\beta$ & (0,1) & (0,10) & (0,10) \\ \hline
\end{tabular}}
\end{table}

In CIFAR-10 dataset and CIFAR-100 dataset, we use the modification matrix $P_{ds}$ to modify the weight gradient of the three CNN layers, to finish the learning of new tasks. Meanwhile, as stated in Section \ref{sect:theproposedenhancedowm}, using $P_{ds}$ to modify the weight gradient $\Delta W_n$ will further increase the impact of new knowledge on the neural network. The authors in \cite{r28} give a point that the last fully connected layer has a strong bias towards new classes, so we use $P_s$ to modify the gradient of the weight of three fully connected layers.

\subsubsection{Evaluation Metrics:}In general, the performance of alleviating forgetting is evaluated by \textit{Average Accuracy} (AA for short), which is a average accuracy of classification over all sequentially learned tasks.
\begin{equation}
    \label{eq_12}
    AA=\frac{1}{N}\sum\nolimits_{i=1}^{N}acc_{i,N}
\end{equation}
where $N$ is the total number of learned tasks and $acc_{i,N}$ is the accuracy of $i$-th learned task.

It is worth noting that two more reasonable metrics BWT \cite{r29} and FM \cite{r30} for evaluating CF have been presented in resent years, as they adopt the mean difference between $acc_{i,i}$ and $acc_{i,N}$ as the evaluation against CF other than AA with $acc_{i,N}$. The metrics BWT and FM are as follows.

\begin{equation}
    \label{eq_14}
    BWT=\frac{1}{N-1} \sum\nolimits_{i=1}^{N} acc_{i,N}-acc_{i,i}
\end{equation}
\begin{equation}
    \label{eq_15}
    FM=\frac{1}{N-1} \sum\nolimits_{i=1}^{N-1} \underset{t \in \{1,...,N-1\}}{max}(acc_{i,t}-acc_{i,N})
\end{equation}

However, we discover the weakness of the BWT and FM, that is, when the ACC values of learned tasks are lower, the values of BWT and FM are also  small, which results in a false result, i.e., the algorithm has a good performance against CF. To overcome this weakness, we propose a more reasonable metric, namely MRR, which uses the quotient of $acc_{i,i}$ and $acc_{i,N}$ to avoid the weakness. The proposed MRR is defined as follows.

\begin{equation}
    \label{eq_13}
    MRR=\frac{1}{N-1} \sum\nolimits_{i=1}^{N-1} \frac{acc_{i,N}}{acc_{i,max}}
\end{equation}
where $acc_{i,max}$ is the maximum accuracy of the $i$-th task in the whole training process. Note that, MRR $\in [0,1]$. A larger MRR indicates a better effect of alleviating forgetting.

\begin{center}
\begin{table*}[t]
\setlength{\abovecaptionskip}{1pt}\centering
\caption{AA and MRR over all tasks in a sequence after the learning of all tasks is completed} \label{tab4_2}
\resizebox{1.0\textwidth}{28mm}{
\begin{tabular}{|c|c|cccccccc|}
\hline
Metric  & Model & \begin{tabular}[c]{@{}c@{}}Shuffled MNIST\\ (3 tasks)\end{tabular} & \begin{tabular}[c]{@{}c@{}}Shuffled MNIST\\ (10 tasks)\end{tabular} & \begin{tabular}[c]{@{}c@{}}Shuffled MNIST\\ (20 tasks)\end{tabular} & \begin{tabular}[c]{@{}c@{}}CIFAR-10\\ (5 tasks)\end{tabular} & \begin{tabular}[c]{@{}c@{}}CIFAR-100\\ (2 tasks)\end{tabular} & \begin{tabular}[c]{@{}c@{}}CIFAR-100\\ (5 tasks)\end{tabular} & \begin{tabular}[c]{@{}c@{}}CIFAR-100\\ (10 tasks)\end{tabular} & \begin{tabular}[c]{@{}c@{}}CIFAR-100\\ (20 tasks)\end{tabular} \\ \hline
\multirow{6}{*}{AA} & EWC & 0.9427 & 0.8820 & 0.6860 & 0.1879 & 0.2646  & 0.1325  & 0.0765 & 0.0425 \\
  & PGMA & 0.9814 & 0.8895 & 0.7010 & 0.4111 & 0.3496 & 0.3001 & 0.2195  & 0.1791  \\
  & OGD  & 0.9380 & 0.8681 & 0.7485 & 0.3466 & 0.3925 & 0.3062 & 0.2074   & 0.1458   \\
  & OGD+ & 0.9403 & 0.9047 & 0.7874 & 0.3785 & 0.4224 & 0.3127 & 0.3060   & 0.1737   \\
  & OWM  & \textbf{0.9834} & 0.9458 & 0.8858 & 0.5316 & 0.4066 & 0.3504 & 0.3149  & 0.2774  \\
  & ER-MIR & 0.9067 & 0.7921 & 0.7594 & 0.4249 &0.3152 & 0.2148 & 0.2363 &0.1935 \\
  & AE-MIR & 0.9129 &0.8433 & 0.7438 &0.3502 &0.3611 & 0.1118  & 0.1066 &0.0590 \\
  & \textbf{Our EOWM} & 0.9832 & \textbf{0.9516} & \textbf{0.8923} & \textbf{0.8223} & \textbf{0.5535} & \textbf{0.5068} & \textbf{0.4328}  & \textbf{0.3782}  \\ \hline
\multirow{6}{*}{MRR} & EWC  & 0.9639  & 0.9203 & 0.7134 & 0 & 0.0087 & 0.0010  & 0 & 0 \\
  & PGMA & 0.9956  & 0.9246 & 0.7292 & 0.4799  & 0.6348 & 0.6540  & 0.4466 & 0.2549 \\
  & OGD  & 0.9874  & 0.9431 & 0.7680 & 0.3800 & 0.7529 & 0.4417  & 0.2867 & 0.1752 \\
  & OGD+ & 0.9894  & 0.9408 & 0.8111 & 0.4110 & 0.7913 & 0.4602  & 0.4019 & 0.2082 \\
  & OWM  & 0.9950  & 0.9711 & 0.9132 & 0.7083  & 0.5891 & 0.6834  & 0.6523 & 0.6606 \\
  & ER-MIR & 0.9694 & 0.9137 &0.8855 &0.7263 &0.6375 &0.6293 &0.4654 &0.3234 \\
  & AE-MIR & 0.9931 & 0.9219 &0.8189 &0.7263 &0.8636 &0.6448 &0.5031 & 0.1816\\
  & \textbf{Our EOWM} & \textbf{0.9965}  & \textbf{0.9752} & \textbf{0.9213} & \textbf{0.8649}  & \textbf{0.8715} & \textbf{0.7294}  & \textbf{0.8154} & \textbf{0.7285} \\ \hline
\end{tabular}}
\end{table*}
\end{center} 
\subsection{Experimental results}
\subsubsection{Performance of alleviating forgetting:}\begin{figure*}[!t]
 \subfigure[]{
  \includegraphics[scale=0.24]{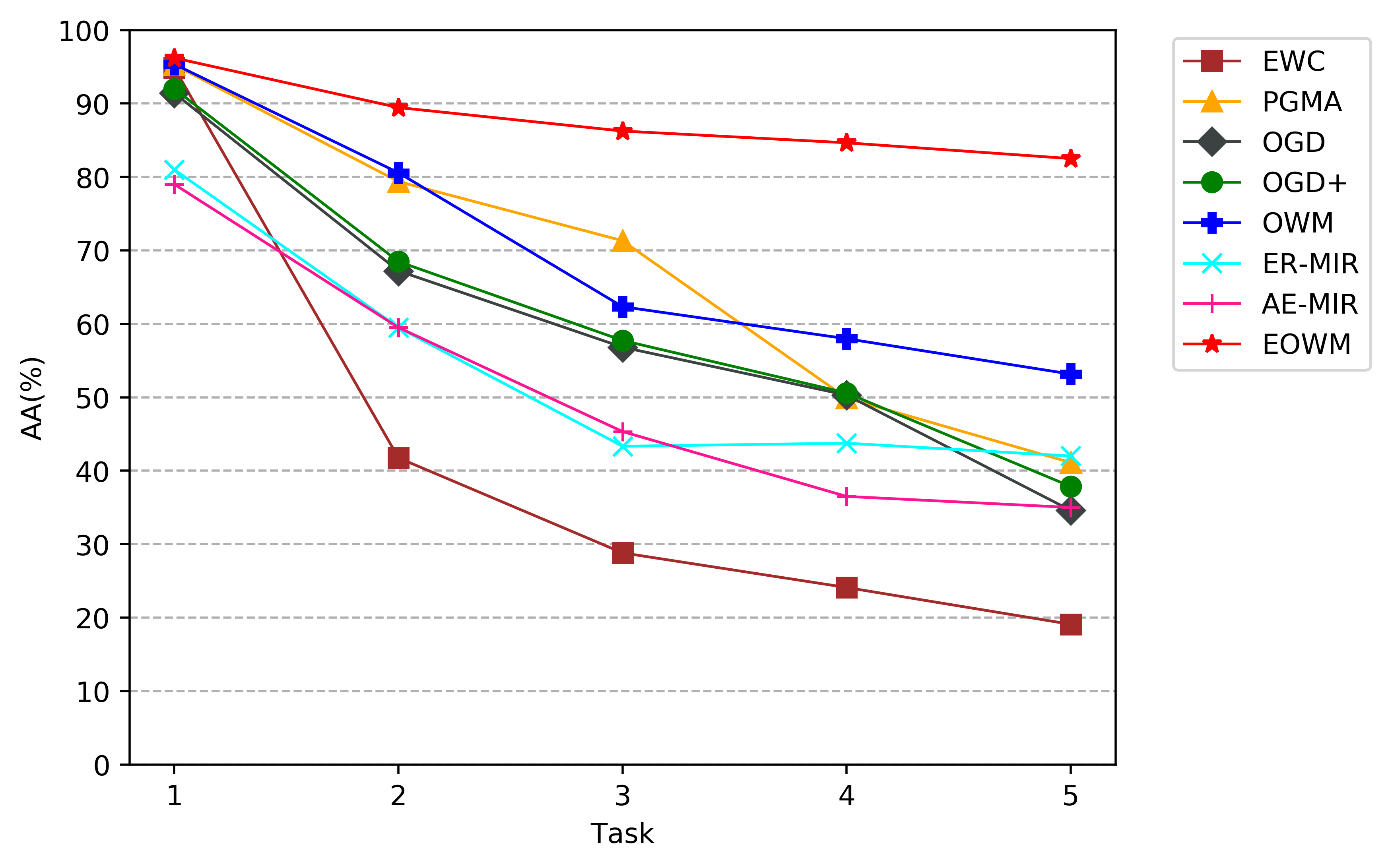}}
 \subfigure[]{
  \includegraphics[scale=0.235]{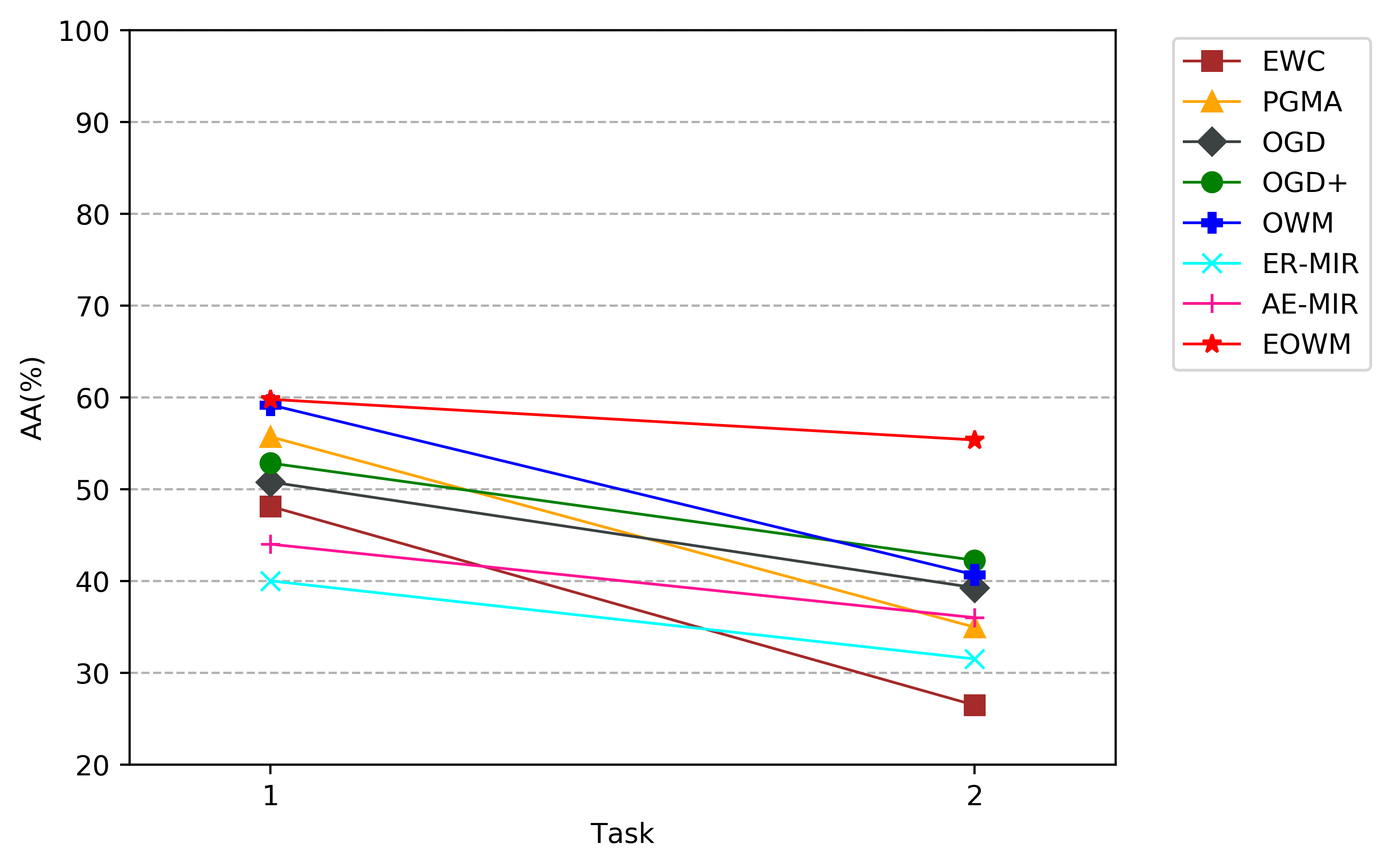}}
 \subfigure[]{
  \includegraphics[scale=0.235]{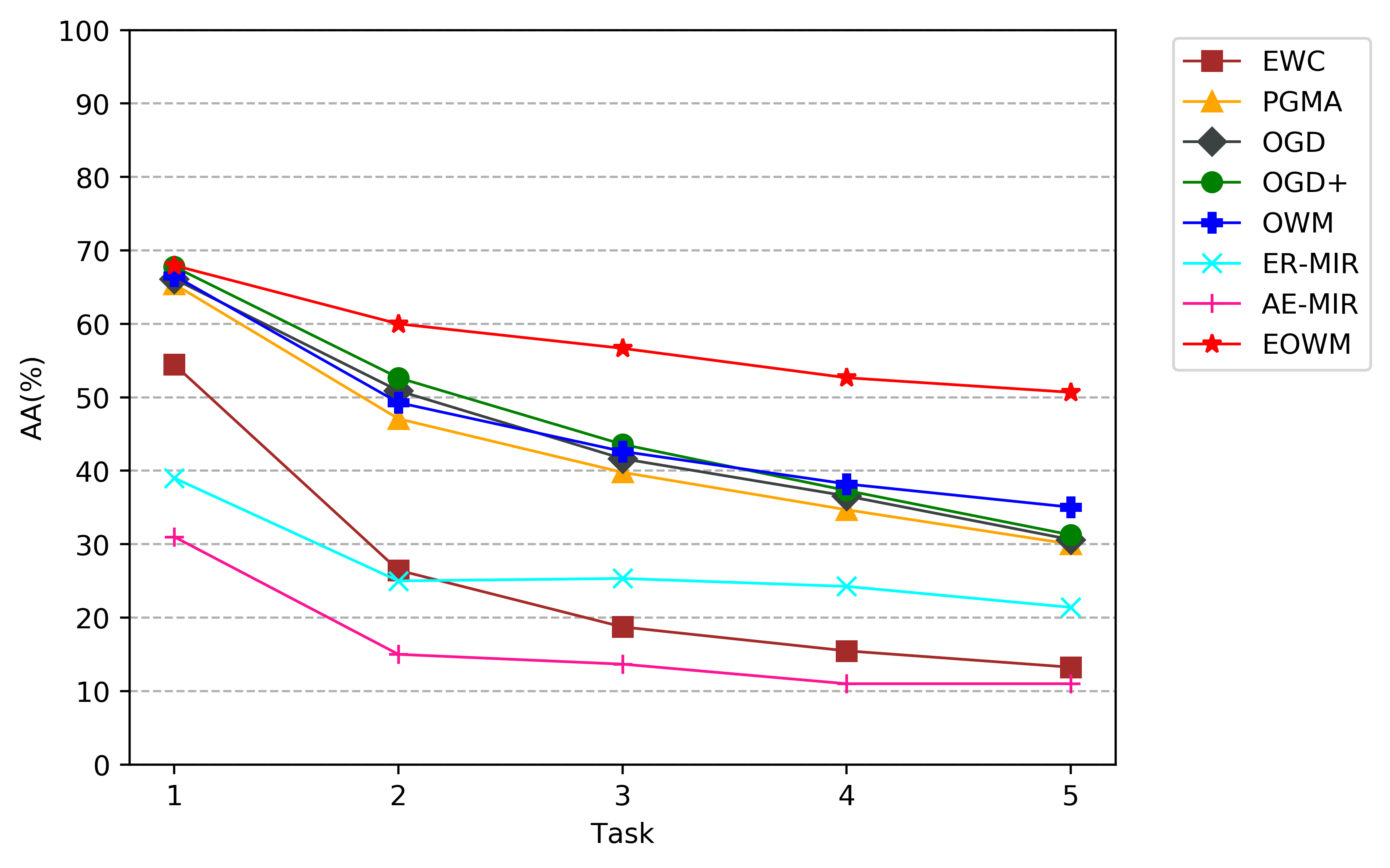}}
 \subfigure[]{
  \includegraphics[scale=0.24]{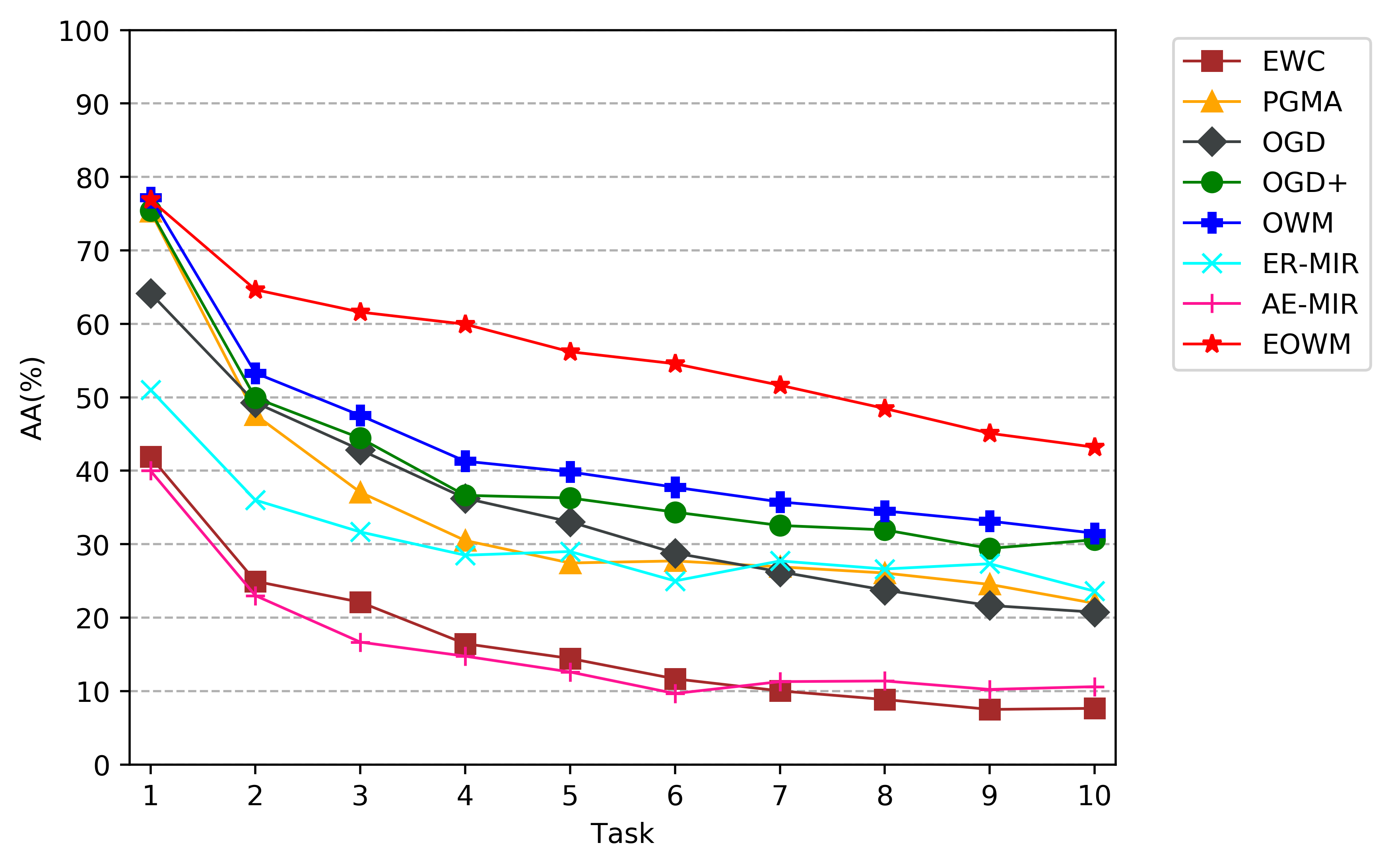}}
 \subfigure[]{
  \includegraphics[scale=0.24]{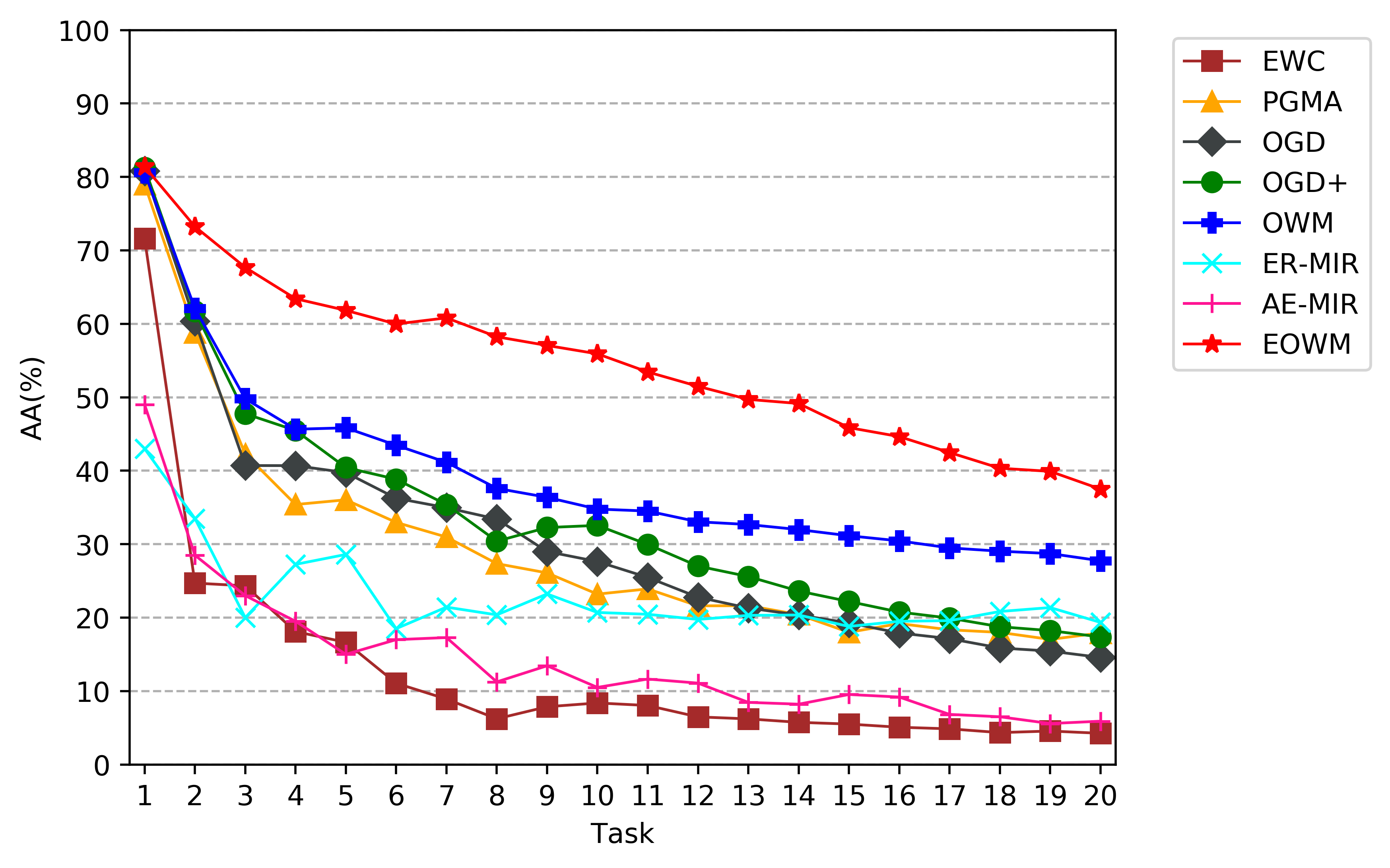}}
 \subfigure[]{
  \includegraphics[scale=0.24]{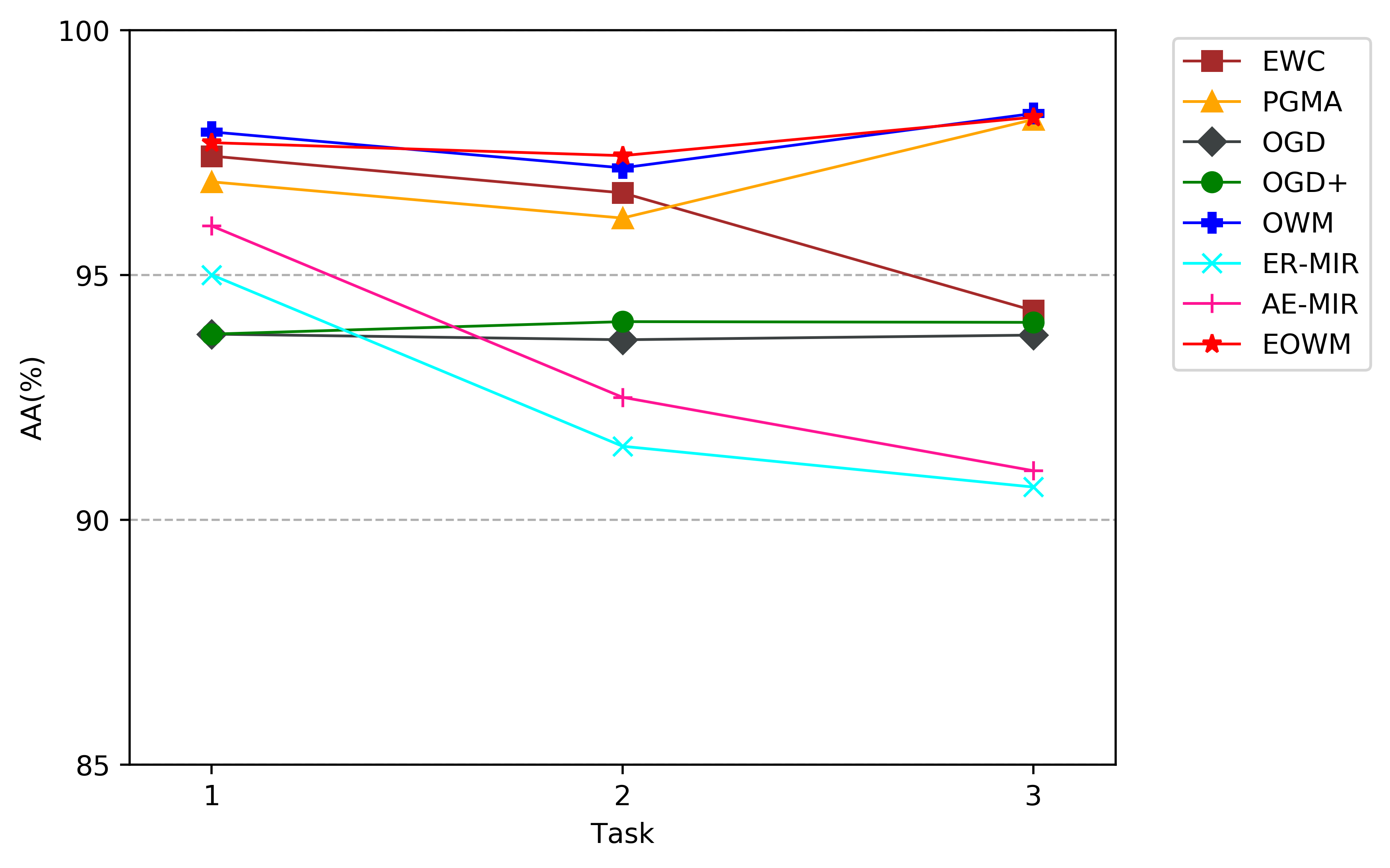}}
 \subfigure[]{
  \includegraphics[scale=0.24]{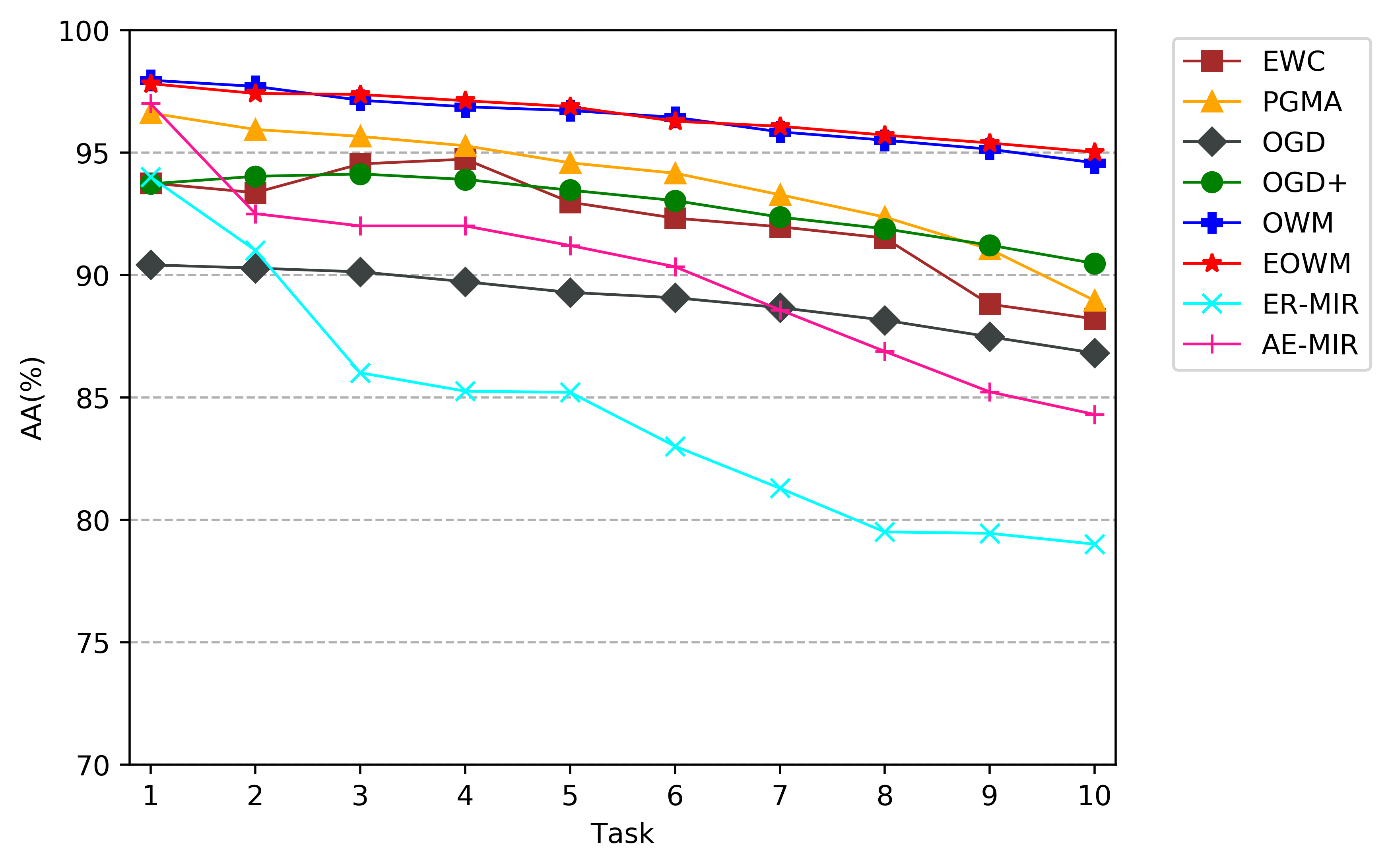}}
 \subfigure[]{
  \includegraphics[scale=0.25]{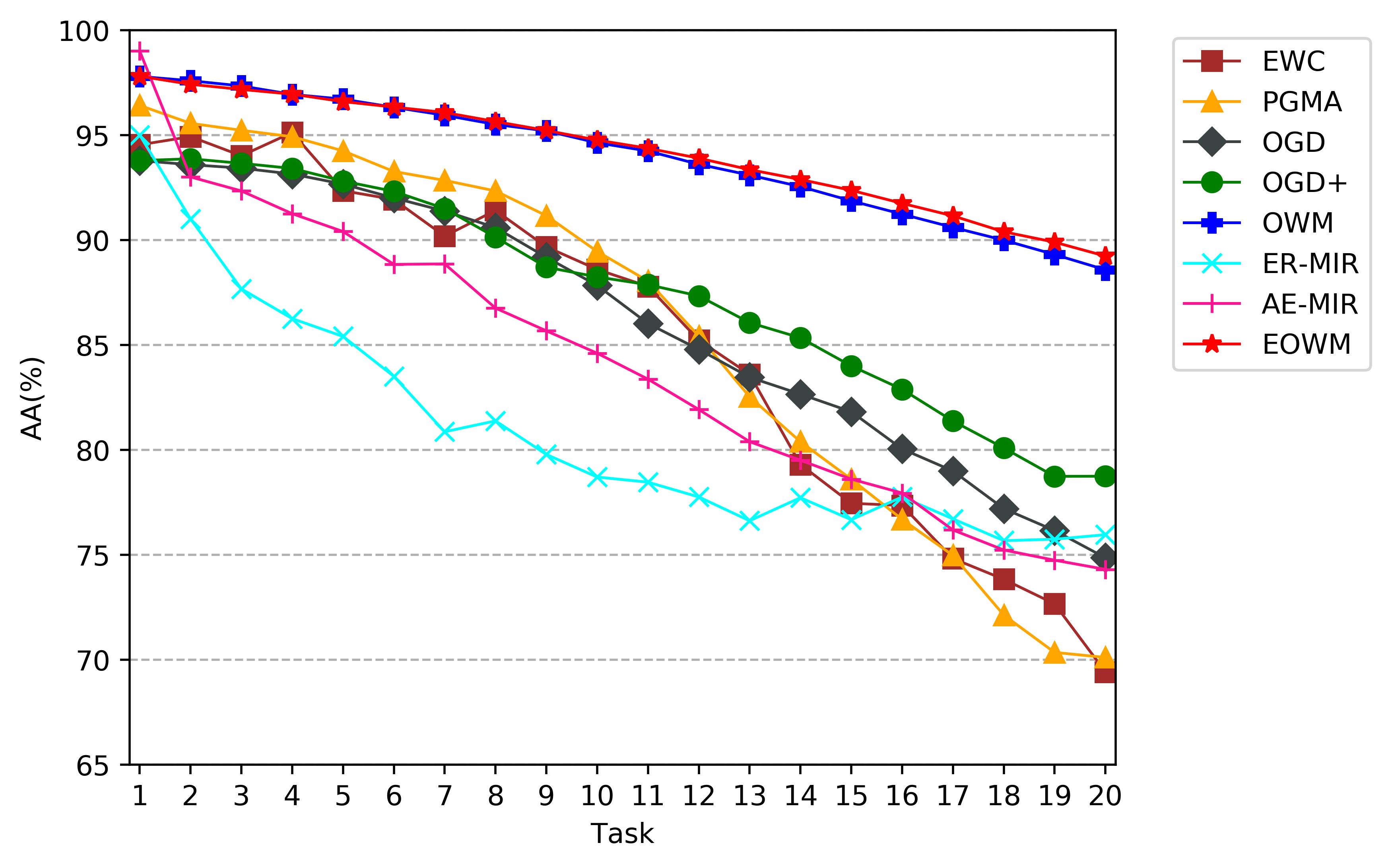}}
 \caption{The average accuracy of EOWM and five baselines on Shuffled MNIST, CIFAR-10, and CIFAR-100. (a) CIFAR-10 with 5 tasks, (b) CIFAR-100 with 2 tasks, (c) CIFAR-100 with 5 tasks, (d) CIFAR-100 with 10 tasks, (e) CIFAR-100 with 20 tasks, (f) Shuffled MNIST with 3 tasks, (g) Shuffled MNIST with 10 tasks, (h) Shuffled MNIST with 20 tasks. }
 \label{fig_3}
 \end{figure*}
In this subsection, we report our evaluation results (the AA and MRR are shown in Table \ref{tab4_2} and Fig. \ref{fig_3}) of the proposed EOWM and other five baselines in two cases (tasks are similar or dissimilar).
From Table \ref{tab4_2} and Fig. \ref{fig_3}, we can draw the following conclusions.

1) When tasks are similar, the EOWM is consistently superior to other algorithms except in the setting that has three tasks on Shuffled MNIST dataset. In the other three settings on Shuffled MNIST dataset, the difference between the EOWM and the OWM is -0.0002, 0.0058 and 0.0065  with the AA metric, respectively. It indicates that in the case of similar tasks, compared to the OWM, as the number of tasks increases, the EOWM has more advantages in learning the optimal weight of a new task.

2) The EOWM has the highest MRR performance on Shuffled MNIST dataset in the case of similar tasks. The difference between our method and OWM is 0.0015, 0.0041 and 0.0081,  respectively. Such a result means that EOWM performs better in alleviating forgetting with the increasing number of tasks when tasks are similar compared to the OWM. Therefore, EOWM has higher potential in finishing continuous learning. 

3) When tasks are dissimilar, our EOWM gains the highest accuracy in all settings on CIFAR-10 and CIFAR-100, which means that our EOWM has significant advantages in learning new features of new tasks. Moreover, in the case of dissimilar tasks, our EOWM yields the best MRR, which indicates that properly enlarge the influence of new task does not have a negative impact on alleviating forgetting.
\begin{table}[h]
\setlength{\abovecaptionskip}{5pt}\centering\caption{The  performance of the AA and MRR over 2 tasks under different combinations of $w_1$ and $w_2$ }\label{tab4_3}
\resizebox{0.5\textwidth}{15mm}{
\begin{tabular}{|c|c|c|c|c|}
\hline
\multirow{2}{*}{Metric} & \multicolumn{4}{c|}{Shuffled MNIST(2 Task)}               \\ \cline{2-5} 
& Coefficient settings           & $w_2=0.05$ & $w_2=0.1$ & $w_2=0.15$ \\ \hline
AA & \multirow{2}{*}{$(w_1+w_2)$=1} & 97.41  & 97.45  & 97.56  \\ \cline{1-1} \cline{3-5} 
MRR &                                & 0.9938 & 0.9964 & 0.9983 \\ \hline
\multirow{2}{*}{Metric} & \multicolumn{4}{c|}{MNIST(2 Task)}               \\ \cline{2-5} 
& Coefficient settings           & $w_2=0.05$ & $w_2=0.1$ & $w_2=0.15$ \\ \hline
AA & \multirow{2}{*}{$(w_1+w_2)$=1}& 87.16  & 87.43 & 87.77  \\ \cline{1-1} \cline{3-5} 
MRR&                                & 0.7742 & 0.7780& 0.7871 \\ \hline
\end{tabular}}
\end{table}
\subsubsection{Ablation study:}In this part, we evaluate the effect of the size of the projection of gradient on $\Omega _{n-1}$ and $\Omega _{n-1}^{\perp}$ on the ability of the neural network to learn new knowledge and alleviate forgetting. We fix $w_1$ and set $w_2$ to different values to adjust the influences of the two gradient projections. We evaluate the impact both using the AA and MRR metrics. When tasks are similar, we shuffle the MNIST and construct two joint tasks, where both have ten classes. However, when tasks are dissimilar, we divide the  MNIST dataset into two disjoint tasks, where both have five classes. The neural network consists of two fully connected layers.

As shown in Table \ref{tab4_3}, when tasks are similar, fixing the value of $w_1+w_2$, both the AA and MRR of the neural network will increase. The reason is that if we fix $w_1+w_2$, enlarging $w_2$ will decrease $w_1$ which means increase the impact of old knowledge and decrease the impact of new knowledge. So the MRR increases and the neural network saves more old knowledge and the AA increases.

When tasks are dissimilar, fixing the value of $w_1+w_2$, the AA increase and the MRR first rises and then decreases. The reason is that if fixing the value of $w_1+w_2$,  a larger $w_2$ indicates a smaller $w_1$, which means increase the impact of old knowledge and decrease the impact of new knowledge.As the increase of $w_2$, the MRR first rises because of the new knowledge's influence less than the old knowledge's and finally decrease because of the new knowledge has great influence when new tasks learning. When the neural network forgets less than it learns, the AA will increase. When the neural network forgets less than it learns, the AA will increase.

\section{Conclusion}
In this study, we first theoretically study the essential weaknesses of the existing OWM methods. Then we reveal and prove the facts that: 1) none of the existing $P_{OWM}$ of the OWM methods take advantage of all the necessary knowledge, and 2) the existing methods rely on an unpractical assumption that the classes of sequential tasks are disjoint. We propose a new enhanced projection operator, i.e., $P_{EOWM}$, which takes into account all the knowledge in $P_{OWM}$ and removes the unrealistic assumption. On the bases, we propose a new OWM algorithm EOWM followed by introducing a more reasonable metric. Extensive experiments conducted on the benchmarks demonstrate that our EOWM is superior to all of the state-of-the-art continual learning baselines.

\clearpage{\bibliography{references}}
\bibliographystyle{aaai22}

\clearpage{\subsection{Appendix}}
\subsection{A. The proof of Theorem 3}
\textbf{Theorem 3: }According to Eq.\ref{eq_7}, the iterative equation of $P_{EOWM}$ can be obtained as follows.
\begin{equation} 
    P_{EOWM}^{n+1}=
    \begin{cases}
    P_{OWM}^{n+1}(c_1I+c_2Q^{n+1}),\ T_{prev} \simeq T_{n+1} \nonumber \\
    P_{OWM}^{n+1}(c_1I+c_2Q_{ort}^{n+1}),\ T_{prev} \not\simeq T_{n+1}\nonumber \
    \end{cases}
\end{equation}
\textbf{\textit{Proof}: }The derivation process based on the Recursive Least Square(\textit{RLS}) method \cite{r10} is as follows.
According to Eq.\ref{eq_1}, the following formula can be obtained.
\begin{align}
    P_{OWM}^{n+1}&=I-A_{n} \left ( A_{n}^TA_{n}+\alpha I\right)^{-1}A_{n}^T  \nonumber \\
    &=I-\alpha^{-1} A_{n} \left (\alpha^{-1}A_{n}^TA_{n}+ I\right)^{-1}A_{n}^T \nonumber 
\end{align} 

Following equation  can be obtained by Woodbury Matrix Identity\cite{r11}.
\begin{align}
    P_{OWM}^{n+1}=(I+\alpha^{-1}A_{n}A_{n}^T)^{-1} \nonumber
\end{align}

For iterative calculation, the average value $\bar{\textbf{x}}_j$ obtained from the data of each task $T_j$ is used as the column of the input space, i.e., $A_{n} = (\bar{\textbf{x}}_1,..., \bar{\textbf{x}}_{n})$.
\begin{align}
    P_{OWM}^{n+1}=&(I+\alpha^{-1}A_{n}A_{n}^T)^{-1} = \alpha(\alpha I+A_{n}A_{n}^T)^{-1} \nonumber\\
    =&\alpha(\alpha I+[A_{n-1},\bar{\textbf{x}}_{n}][A_{n-1},\bar{\textbf{x}}_{n}]^T)^{-1} \nonumber\\
    =&\alpha(\alpha I+A_{n-1}A_{n-1}^T+\bar{\textbf{x}}_{n}\bar{\textbf{x}}_{n}^T)^{-1} \nonumber\\
    =&\alpha(\alpha (P_{OWM}^n)^{-1}+\bar{\textbf{x}}_{n}\bar{\textbf{x}}_{n}^T)^{-1}\nonumber
\end{align}

Following equation can be obtained by Matrix Inverse Lemma\cite{r11} .
\begin{align}
P_{OWM}^{n+1}=&\alpha(\alpha^{-1}P_{OWM}^{n}-\frac{\alpha^{-1}P_{OWM}^n\bar{\textbf{x}}_{n}\alpha^{-1}\bar{\textbf{x}}_{n}^TP_{OWM}^n}{1+\alpha^{-1}\bar{\textbf{x}}_{n}^T
    P_{OWM}^n\bar{\textbf{x}}_{n}})\nonumber
\end{align}

Then,
\begin{align}
    P_{OWM}^{n+1}=&P_{OWM}^{n}-\frac{P_{OWM}^n\bar{\textbf{x}}_{n}}{\alpha+\bar{\textbf{x}}_{n}^T
    P_{OWM}^n\bar{\textbf{x}}_{n}}\bar{\textbf{x}}_{n}^TP_{OWM}^n\nonumber\\
    =&P_{OWM}^{n}-\boldsymbol{\kappa}_x\bar{\textbf{x}}_n^TP_{OWM}^{n}  \nonumber
    \nonumber 
\end{align}

Similarly, let the average value $\overline{\textbf{W}}_j$ obtained from the weights after each task $T_j$ trained as the column of the weight space, i.e., $W_{n} = (\overline{\textbf{W}}_1,..., \overline{\textbf{W}}_{n})$. Then we can directly obtain the iterative formulas of $Q_{ort}$ and $Q$ as follows.
\begin{equation}
\ \ \ \ \ \ \ \ \ \ \ \ \ \ \ \ \ \ \ \ \ \ \ \ \ \ \ \ \ \
\begin{cases}
Q_{ort}^{n+1} &= Q_{ort}^{n} - \boldsymbol{\kappa}_{W}  \overline{\textbf{W}}_{n}^TQ_{ort}^{n} \nonumber\ \\ 
Q^{n+1} &= I-Q_{ort}^n \nonumber\ \ \ \ \ \ \ \ \ \ \ \ \ \ \ \ \ \ \ \ \ \ \ \ \ \ \ \ \ \ \ \ \ \ \ \ \  \hfill\blacksquare
\end{cases}
\end{equation}

\subsection{B. The proof of Theorem 4}
\textbf{Theorem 4: }The time complexity of the EOWM algorithm is  $O(N_c^3N_n)$, where $N_c$ and $N_n$ are the number of the columns of the weight and the number of neurons of a layer, respectively(proof shown in Appedix).

\textbf{\textit{Proof: }}
For the learning of any layer $L_l$ of neural network, the time complexity of calculating $P_{OWM}$ is $O(N_c^2)$ for each iteration\cite{r10}, where $N_c$ is the number of columns of the weight of the layer $L_l$.

The calculation process of $P_{EOWM}$ shown in Eq.\ref{eq_9} includes one matrix addition, one matrix multiplication and two iterations to calculate the projection operator.
 
According to the process of matrix operation, matrix multiplication has the highest time complexity $O(N_c^3)$. Therefore, the time complexity of calculating $P_{EOWM}$ in each iteration is $O(N_c^3)$.
 
To sum up, when the network layer $L_l$ is trained with the EOWM algorithm, the time complexity is $O(N_c^3N_n)$, where $N_n$ is the number of neurons of the layer $L_l$. $\hfill\blacksquare$

\subsection{C. The proof of Theorem 5} 
\textbf{Theorem 5 :} The upper bound of the minimum number of learnable tasks of our proposed EOWM is $rank(P_{EOWM})$, which is fomulated as follows. 
\begin{equation} 
\footnotesize{
    rank(P_{EOWM})<=
    \begin{cases}
    Rank_{OWM}+rank(P_{OWM}Q),\ T_{prev} \simeq T_{n+1} \nonumber\ \\
    Rank_{OWM}+rank(P_{OWM}Q_{ort}),\ T_{prev} \not\simeq T_{n+1} \nonumber\
    \end{cases}
    }
\end{equation}
where $Rank_{OWM}$ is the maximum rank of $P_{OWM}$ when $rank(P_{OWM}Q)>0$ and $rank(P_{OWM}Q_{ort})>0$.

\textbf{\textit{Proof: }}According to Eq.\ref{eq_7}, when new and old tasks are not similar, the rank of $P_{EOWM}$ can be calculated as follows.
\begin{equation}
\small{
\begin{aligned}
    rank(P_{EOWM})&=rank(P_{OWM}(c_1I+c_2Q_{ort}))\\ 
    &=rank(P_{OWM}+P_{OWM}Q_{ort}))\\ 
    &<=rank(P_{OWM})+rank(P_{OWM}Q_{ort}) \nonumber
\end{aligned}
}
\end{equation}

According to the analysis of \cite{r8}, when $\alpha=0$, with the continuous learning of new tasks, the neural network corrected by $P_{OWM}$ will reach its limits, in other words, the rank of $P_{OWM}$ reaches the maximum value recorded as $Rank_{OWM}$, i.e.,
\begin{equation}
rank(P_{OWM})<=Rank_{OWM} \nonumber
\end{equation}

From the properties of projection operators, $QQ_{ort} = 0$ and $P_{OWM}Q_{ort} \neq 0$, so $rank(P_{OWM}Q_{ort})>0$. 
 
When new and old tasks are similar, the same conclusion can be drawn, which will not be repeated here.$\hfill\blacksquare$

\end{document}